\documentclass[journal]{IEEEtran}
\usepackage{amsmath,amsfonts}
\usepackage{algorithmic}
\usepackage{algorithm}
\usepackage{array}
\usepackage{textcomp}
\usepackage{stfloats}
\usepackage{url}
\usepackage{verbatim}
\usepackage{graphicx}
\usepackage{caption}
\usepackage{subcaption}
\usepackage{cite}
\usepackage{booktabs}
\usepackage{multirow}
\usepackage{bbding}
\usepackage{xcolor}
\begin{document}

\title{Linguistic Knowledge Transfer Learning for Speech Enhancement}

\author{Kuo-Hsuan Hung,~\IEEEmembership{Student Member,~IEEE,} Xugang Lu, Szu-Wei Fu, Huan-Hsin Tseng,
\par
Hsin-Yi Lin, Chii-Wann Lin, and Yu Tsao,~\IEEEmembership{Senior Member,~IEEE}

\thanks{Kuo-Hsuan Hung and Chii-Wann Lin are with the Department of Biomedical Engineering, National Taiwan University, Taipei, Taiwan (email: d07528023@ntu.edu.tw; cwlinx@ntu.edu.tw).}
\thanks{Xugang Lu is with National Institute of Information and Communications Technology, Japan (email: xugang.lu@nict.go.jp).}
\thanks{Szu-Wei Fu is with NVIDIA, Taipei, Taiwan (email: szuweif@nvidia.com).}
\thanks{Huan-Hsin Tseng is with the AI \& ML Department, Brookhaven National Laboratory, Upton NY, USA (email: htseng@bnl.gov).}
\thanks{Hsin-Yi Lin is with the Department of Mathematics and Computer Science, Seton Hall University, South Orange, NJ 07079, USA (email: hsinyi.lin@shu.edu).}
\thanks{Yu Tsao is with the Research Center for Information Technology Innovation, Academia Sinica, Taipei, Taiwan (e-mail: yu.tsao@sinica.edu.tw).}
}


\markboth{Journal of \LaTeX\ Class Files,~Vol.~14, No.~8, August~2021}%
{Shell \MakeLowercase{\textit{et al.}}: A Sample Article Using IEEEtran.cls for IEEE Journals}


\maketitle

\begin{abstract}
Linguistic knowledge plays a crucial role in spoken language comprehension. It provides essential semantic and syntactic context for speech perception in noisy environments. However, most speech enhancement (SE) methods predominantly rely on acoustic features to learn the mapping relationship between noisy and clean speech, with limited exploration of linguistic integration. While text-informed SE approaches have been investigated, they often require explicit speech-text alignment or externally provided textual data, constraining their practicality in real-world scenarios. Additionally, using text as input poses challenges in aligning linguistic and acoustic representations due to their inherent differences. In this study, we propose the Cross-Modality Knowledge Transfer (CMKT) learning framework, which leverages pre-trained large language models (LLMs) to infuse linguistic knowledge into SE models without requiring text input or LLMs during inference. Furthermore, we introduce a misalignment strategy to improve knowledge transfer. This strategy applies controlled temporal shifts, encouraging the model to learn more robust representations.
Experimental evaluations demonstrate that CMKT consistently outperforms baseline models across various SE architectures and LLM embeddings, highlighting its adaptability to different configurations. Additionally, results on Mandarin and English datasets confirm its effectiveness across diverse linguistic conditions, further validating its robustness. Moreover, CMKT remains effective even in scenarios without textual data, underscoring its practicality for real-world applications. By bridging the gap between linguistic and acoustic modalities, CMKT offers a scalable and innovative solution for integrating linguistic knowledge into SE models, leading to substantial improvements in both intelligibility and enhancement performance.
\end{abstract}


\section{Introduction}
\IEEEPARstart{S}{peech} enhancement (SE) aims to improve the quality and intelligibility of speech signals that have been degraded by environmental factors such as background noise, reverberation, and channel distortion \cite{fu2022uformer, cauchi2015combination}. It is widely used as a front-end processing technique in various real-world applications, including speaker recognition \cite{sadjadi2010assessment, Michelsanti2017, mowlaee2012joint}, automatic speech recognition (ASR) \cite{Li2014AnOO, weninger2015speech} and hearing aid \cite{doclo2018binaural}.
Traditional SE methods can be broadly classified into statistical-based approaches and signal modeling-based approaches. Statistical-based methods, such as spectral subtraction \cite{berouti1979enhancement, tanaka2013hybrid} and Wiener filtering \cite{scalart1996speech, abd2014speech, manamperi2022gmm}, estimate noise characteristics and apply suppression filters to attenuate noise components in the spectral domain. In contrast, signal modeling-based methods, including subspace algorithms \cite{dendrinos1991speech, ephraim1995signal}, harmonic models \cite{frazier1976enhancement}, and linear prediction techniques \cite{atal1979predictive, borgstrom2012linear}, rely on assumptions about the intrinsic structures of speech and noise signals to enhance speech quality. However, these conventional approaches often require prior knowledge of noise characteristics and struggle to handle non-stationary noise conditions.

With the rapid advancements of deep learning (DL) algorithms, deep neural network (DNN)-based models have been widely adopted to SE. These models have evolved from simple feed-forward networks \cite{lu2013speech, koizumi2021df, delcroix2017multichannel, furnon2021dnn, kang2018dnn} to more advanced architectures, including convolutional neural networks (CNNs) \cite{park2016fully, pandey2019new}, recurrent neural networks (RNNs) \cite{chen2015integration, strake2020speech}, and hybrid models that combine both \cite{hu2020dccrn, chakrabarty2019time}. More recently, Transformer-based \cite{wang2021tstnn, deoliveira22_interspeech} and Conformer-based \cite{abdulatif2024cmgan, lu2023mp} architectures have demonstrated superior performance by effectively capturing both local and global dependencies in speech signals. Meanwhile, generative models such as generative adversarial networks (GANs) \cite{pascual17_interspeech, fu2019metricgan} and diffusion-based models \cite{welker22speech, lemercier2023storm} have emerged as powerful alternatives, leveraging data-driven approaches to learn complex signal distributions for high-quality SE. 
In addition to approaches that rely solely on acoustic information, SE research has increasingly explored multimodal techniques that integrate auxiliary information to enhance system performance. For example, incorporating visual cues from lip movements or facial expressions \cite{gabbay2017visual, ephrat2018looking} has been shown to significantly improve model effectiveness, particularly in low SNR environments. However, visual-based approaches are sensitive to variations in lighting conditions, camera angles, and occlusions. In contrast, articulatory movement features, which directly capture speech-related motions within the vocal tract, are more robust to environmental changes. Recent studies have demonstrated that integrating articulatory movements with acoustic signals can further improve SE performance \cite{chen2021study, wang2022multi}.

In addition to visual cues and articulatory movements, several approaches have been explored to incorporate phonetic information into SE. Early studies proposed spectral feature mapping techniques, in which phoneme-level classification loss was introduced as an auxiliary objective to improve phonetic consistency \cite{bagchi2018spectral}. Subsequent research investigated perceptual loss methods, utilizing pre-trained phoneme classifiers to guide SE models, ensuring that phonetic structures were better preserved \cite{kataria2021perceptual, plantinga2021perceptual}. Building on these foundations, symbolic sequential modeling techniques were developed to encode speech signals as discrete phoneme-like representations via vector quantization, allowing SE models to capture phonetic structures more effectively \cite{liao19_interspeech}. Another line of work explored the use of broad phonetic class (BPC) information, incorporating auxiliary loss functions derived from ASR models to guide SE training and improve enhancement performance \cite{lu2023improving}. These phonetic-aware approaches provide meaningful subword-level representations, enabling SE models better capture phonetic details and improve overall performance. However, they remain constrained in incorporating high-level semantic information, which is crucial for speech comprehension.

Linguistic information is essential in speech processing, as it provides a higher-level semantic and syntactic context that enhances speech understanding. Previous studies has demonstrated that language familiarity can influence auditory attention, allowing listeners to concentrate on speech even in noisy environments \cite{barenholtz2016language}. This phenomenon highlights the role of linguistic information in shaping speech perception, particularly in challenging listening conditions. Similar trends have been observed in SE models, where the language mismatch between training and testing data affects performance \cite{xu2014cross, close2023effect}. These findings indicate the potential for incorporating linguistic information to complement the acoustic characteristics that improve the robustness of SE systems. However, the integration of linguistic knowledge into SE remains relatively unexplored. Given that linguistic information is inherently embedded in textual data, text-informed SE approaches have been explored. Early methods relied on forced alignment between text and speech to enhance the quality of speech signals \cite{kinoshita2015text}, whereas later approaches introduced knowledge distillation techniques to reduce dependency on text during inference \cite{wang2022text}. However, these methods directly incorporate textual inputs without capturing a more comprehensive linguistic representation, making it challenging to effectively extract and integrate linguistic information within the acoustic modality. 

\begin{figure}[tb]
    \centering 
    \begin{subfigure}[b]{0.45\textwidth}
        \centering
        \includegraphics[width=\textwidth]{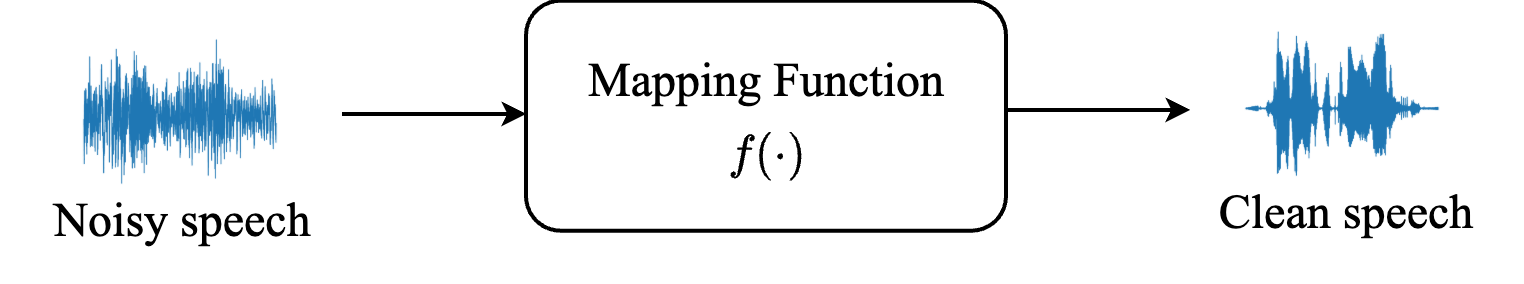}
        \caption{Conventional SE}
        \vspace{0.3cm}
    \end{subfigure}

    \begin{subfigure}[b]{0.45\textwidth}
        \centering
        \includegraphics[width=\textwidth]{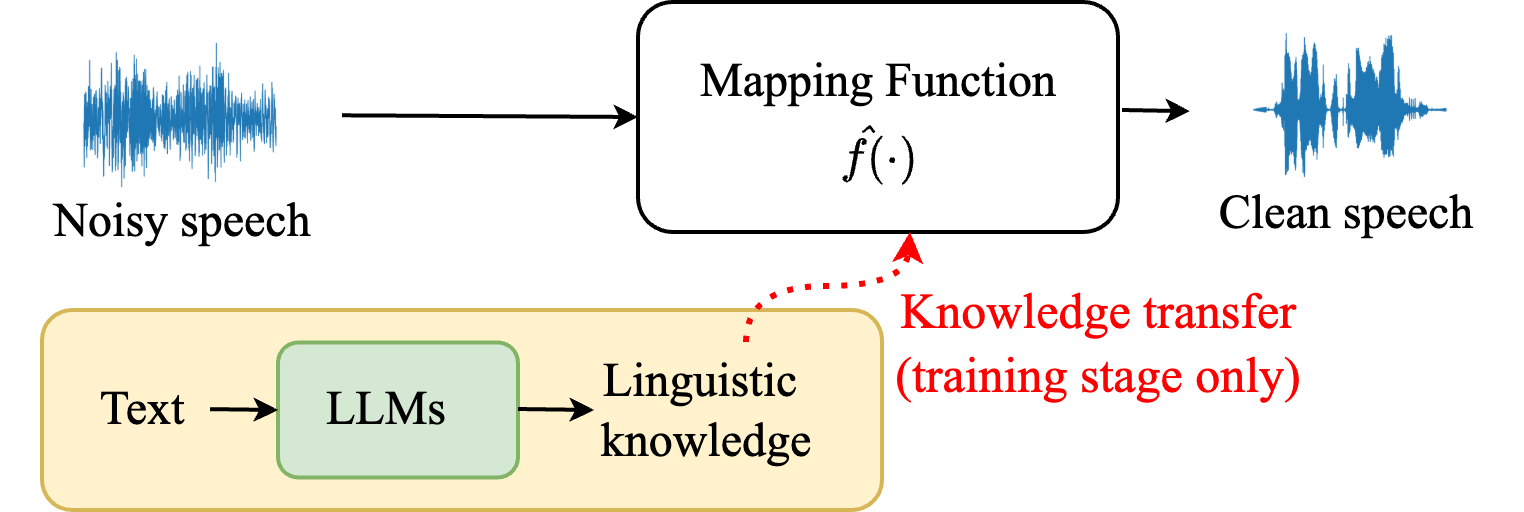}
        \caption{Proposed CMKT learning}
    \end{subfigure}
    \vspace{0.1cm}
    \caption{Comparative flowchart between  conventional SE and the proposed CMKT learning.}
    \label{fig:flow}
\end{figure}

Recently, pre-trained large language models (LLMs) have demonstrated remarkable versatility, providing rich and context-aware linguistic knowledge across diverse tasks \cite{Touvron2023LLaMAOA, huang2023language, zhangetal2023video}. Despite their proven effectiveness in natural language processing tasks, LLMs have not been explored in SE. This study addresses this gap by proposing a novel Cross-Modality Knowledge Transfer (CMKT) learning framework, which leverages LLM embeddings to transfer linguistic knowledge into SE models, as illustrated in Fig. \ref{fig:flow}. The main contributions of this study are as follows: (1) This study is the first to demonstrate the effectiveness of transferring linguistic information derived from LLM to SE models. By integrating LLM embeddings, the CMKT framework significantly improves SE performance across various architectures and configurations. Furthermore, the approach eliminates the need for text data or LLMs during inference, highlighting its practicality in real-world scenarios. 
(2) We propose a misalignment strategy that introduces temporal shifts during the loss function calculation. This approach encourages the model to learn more informative representations by making the prediction task more challenging. Additionally, attention weighting maps are employed to provide interpretability and illustrate the advantages of this strategy. (3) We evaluate our framework on both Mandarin and English datasets, showcasing its effectiveness in diverse linguistic conditions and validating its robustness. (4) Our method remains applicable even in scenarios without text data, demonstrating its practicality to real-world SE tasks where linguistic resources may be limited or inaccurate.

The remainder of this paper is organized as follows. Section \ref{sec:related} reviews the related work on SE and ASR, providing an overview of existing approaches that are relevant to the proposed framework. Section \ref{sec:proposed} introduces the CMKT learning, detailing its methodology and the integration of linguistic and acoustic information. Section \ref{sec:experiments} describes the experimental setup and evaluates the proposed approach under various configurations, demonstrating its effectiveness across different SE models and LLMs. Finally, Section \ref{sec:conclusion} concludes the paper and discusses potential directions for future work.



\section{Related Work}
\label{sec:related}
\subsection{Speech Tasks with Text-based Large Language Models}
Text-based Large Language Models (text-LLMs) have achieved remarkable progress in natural language processing (NLP) \cite{Touvron2023LLaMAOA} and have been expanded to other modalities, including speech and vision \cite{zhangetal2023video}. In speech-related tasks, LLMs are primarily employed in text-oriented applications like ASR \cite{wu2023decoder, rad2023whispering} and text-to-speech (TTS) synthesis \cite{Hao2023BoostingLL, neekhara24_interspeech}. They have also shown potential in less text-dependent tasks, such as speech-emotion recognition (SER) \cite{cheng2024emotionllama, ma2024leveraging} and speech assessment \cite{wang2024enabling}.

One of the tasks most relevant to our work is speech restoration, exemplified by models such as Miipher \cite{koizumi2023miipher}. Miipher leverages w2v-BERT to extract robust speech features and PnG-BERT for linguistic conditioning. This combination effectively addresses challenges such as phoneme masking and deletion caused by noise or compression. By integrating self-supervised speech models with text-based LLMs, Miipher enhances the quality of degraded speech. Although speech restoration and SE are closely related tasks, our approach to SE follows a fundamentally different direction. Unlike previous methods, our framework requires neither corresponding text nor text-based LLMs during inference. This design substantially reduces computational resource requirements while preserving strong SE performance.

Another relevant study is the Timed Text-Regularized Speech Separation (TTR-SS) model \cite{Hsieh2024MultimodalRL}, which is designed for speech separation and aims to disentangle individual audio sources from a mixture. TTR-SS utilizes a multimodal representation loss between timed text and speech to regularize speech separation models. By aligning subword-level text and audio embeddings using pre-trained WavLM and BERT models within a summarizer Transformer framework, this method leverages sentence-level semantics during training. This design not only enhances source separation performance but also removes the need for auxiliary text data during inference. However, TTR-SS rely on explicit alignment between text and speech to guide training. In contrast, our method does not require such alignment annotations, nor does it depend on corresponding text during inference. Furthermore, despite the advancements in text-LLMs across various speech-related tasks, their application to SE remains unexplored.

\subsection{Cross-Modality Knowledge Transfer in Automatic Speech Recognition}
Pre-trained language models (PLMs), such as BERT \cite{devlin2018bert}, encode rich linguistic knowledge that enhances the semantic and contextual understanding of ASR systems \cite{futami20_interspeech, kubo2022knowledge}. Although PLMs are commonly used as external components for post-processing tasks such as rescoring \cite{shin2019effective}, these approaches often incur high computational costs \cite{yu2022non}. Cross-modality knowledge transfer (CMKT) \cite{lu2023cross, lu2024hierarchical, Lu2024TemporalOP} mitigates this issue by embedding linguistic knowledge directly into acoustic models during training. This allows ASR systems to perform independently of PLMs during inference and improves computational efficiency.

Recent advancements in CMKT have introduced effective methods for aligning acoustic and linguistic representations. Optimal Transport (OT) minimizes distributional discrepancies by transforming acoustic features to align with linguistic features encoded by PLMs \cite{lu2023cross}. Hierarchical frameworks further enhance CMKT by embedding linguistic guidance across multiple layers of the acoustic encoder \cite{lu2024hierarchical}. This integration captures both low- and high-level semantic features. Additionally, advanced attention mechanisms, such as Sinkhorn attention, refine cross-modality alignment with iterative normalization, improving the overall knowledge transfer process.

Temporal alignment techniques, such as Temporal Order Preserved OT, address the sequential nature of speech data \cite{Lu2024TemporalOP}. These methods ensure consistency in temporal structures during alignment and preserve the integrity of speech information. By combining these strategies, CMKT enhances the integration of linguistic and acoustic information. It significantly boosts ASR performance without requiring additional computational resources during inference. This makes ASR systems more robust, efficient, and practical for real-world applications.

Although PLMs have successfully helped many tasks in NLP-related tasks with the context-dependent linguistic representation capability, there is few work to show that PLMs could benefit to noisy speech restoration, i.e., SE. As acoustic speech and underlying linguistic content can be regarded as two sides of a coin, it is reasonable to think that transferring linguistic knowledge in acoustic processing could help noisy speech restoration. But whether the PLMs could help in acoustic speech processing for SE, and how to efficiently integrate PLMs for SE is still a challenge. In this study, we propose an efficient model framework to transfer linguistic knowledge from PLMs for SE.

\begin{figure}
\centerline{\includegraphics[width=0.95\columnwidth]{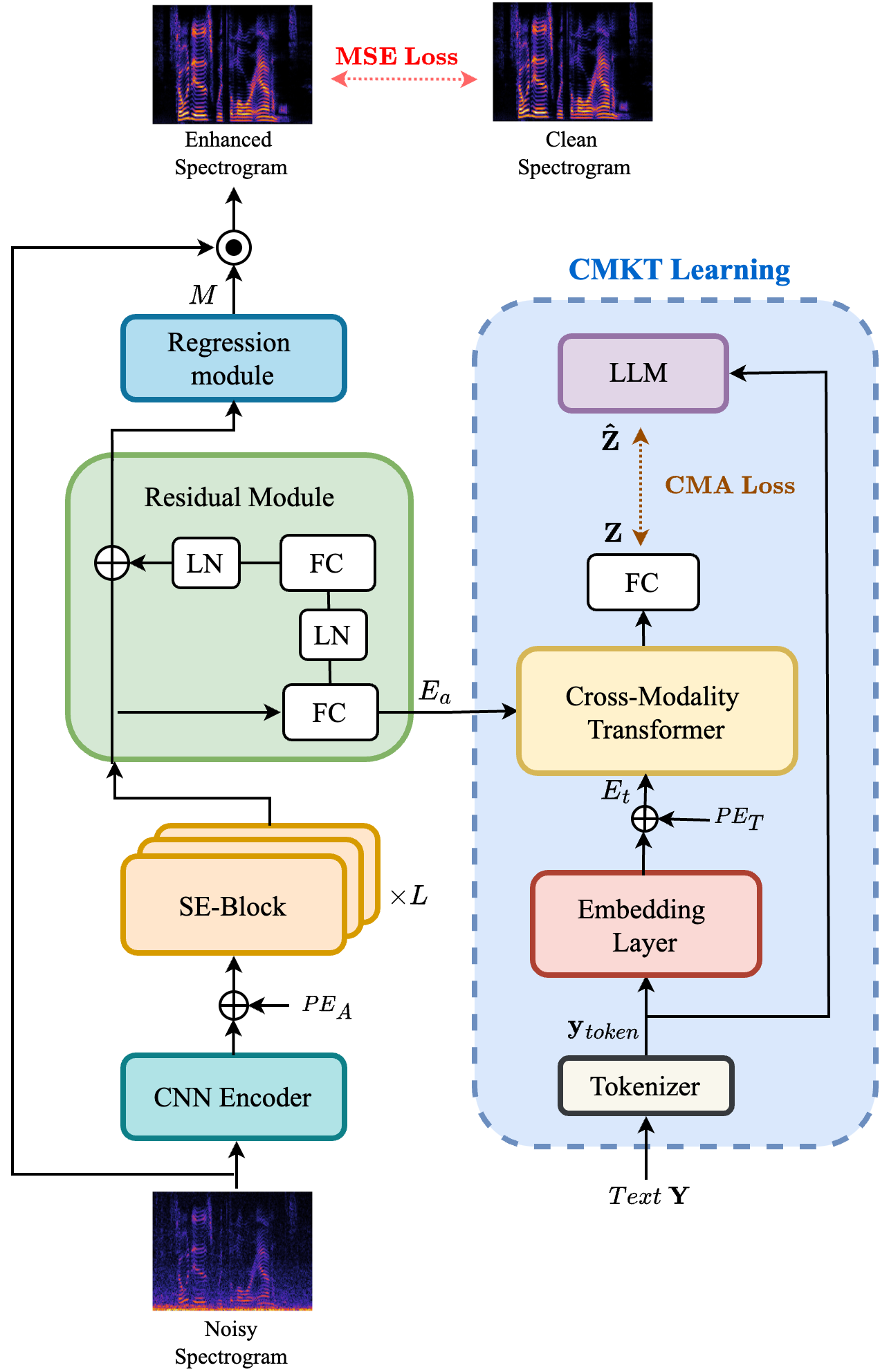}}
\caption{The proposed model architecture diagram. The left branch represents the baseline SE model, while the right branch represents the Text-Speech integration module. Where $\odot$ and $\oplus$ denote as element-wise multiplication and element-wise addition, respectively.
}
\label{fig:model}
\end{figure}

\section{Proposed method}
\label{sec:proposed}

The proposed model framework is illustrated in Fig. \ref{fig:model} and consists of two information processing modalities. The left branch represents the conventional SE model, which operates in the acoustic modality and aims to estimate clean speech from noisy input. The right branch corresponds to the CMKT model, which operates in the text modality and extracts linguistic information from the LLM to enhance acoustic embeddings. Importantly, the right branch is only active during training to optimize SE learning and is not utilized during inference. The following sections provide a detailed description of the model architecture.

\subsection{Speech Enhancement Model}

The SE module is designed to estimate clean speech features from noisy input. Depending on the training objective, SE methods can be broadly classified into mapping-based and masking-based approaches \cite{lee2018phase, nossier2020mapping}. In this study, we adopt a masking-based approach, where the model learns a spectral mask to refine the noisy magnitude spectrogram. Compared to direct mapping, masking-based methods effectively reduce distortion by selectively suppressing noise, while simultaneously preserving speech structure and key spectral components \cite{wang2014training, williamson2015complex}.
The proposed model consists of four key components: a CNN encoder, speech enhancement blocks (SE-blocks), a residual module, and a regression module.

\subsubsection{CNN Encoder}
The CNN encoder receives the noisy log-amplitude spectrogram $\mathbf{X} \in \mathbb{R}^{T_{a} \times F}$ as input for initial acoustic feature extraction, where $T_{a}$ and $F$ represent the time and frequency dimensions of the spectrogram, respectively. The encoder consists of two 2D convolution layers followed by a linear layer. Each convolution layer applies ReLU \cite{agarap2018deep} activation to introduce non-linearity and enhance feature extraction. After processing through the CNN encoder, the extracted speech features are augmented with relative position encoding ($PE_{A}$) to improve generalization across different input lengths. The feature transformation is formulated as follows:

\begin{equation}
\label{eq1}
\begin{aligned}
&\mathbf{H} = \mathrm{CNN\_Encoder} (\mathbf{X})\\
&\mathbf{A}^{in}_{0} = \mathbf{H} + \mathrm{PE}_{A}(\mathbf{H})
\end{aligned}
\end{equation}
Where $\mathbf{H}$ and $\mathbf{A}^{in}_{0}$ represent the output of the CNN encoder and the input to the SE-block, respectively.

\subsubsection{Speech Enhancement Blocks}
\label{sec:seblock}
In this work, we employ the Conformer block \cite{gulati2020conformer} for the SE-blocks, which have proven highly effective in ASR by leveraging the strengths of both Transformers and convolutional neural networks (CNNs). The architecture has also demonstrated notable effectiveness in SE \cite{abdulatif2024cmgan}. Each Conformer block consists of four sequential modules: a feed-forward network ($\mathrm{FFN_1}$) module, a Multi-Head Self-Attention (MHSA) module, a convolution (Conv) module, a layer normalization (LN) and a second feed-forward network ($\mathrm{FFN_2}$) module. The processing in each Conformer block is expressed as follows:
\begin{equation}
\label{eq2}
\begin{aligned}
&\mathbf{A}^{1}_{l} = \mathbf{A}^{in}_{l-1}+ \frac{1}{2}\mathrm{FFN_1}(\mathbf{A}^{in}_{l-1})\\
&\mathbf{A}^{2}_{l} = \mathbf{A}^{1}_{l}+\mathrm{MHSA}(\mathbf{A}^{1}_{l})\\
&\mathbf{A}^{3}_{l} = \mathbf{A}^{2}_{l}+\mathrm{Conv}(\mathbf{A}^{2}_{l})\\
&\mathbf{A}^{4}_{l} = \mathrm{LN}(\mathbf{A}^{3}_{l}+\frac{1}{2}\mathrm{FFN_2}(\mathbf{A}^{3}_{l})),\\
\end{aligned}
\end{equation}
where $l$ ranges from 1 to $L$, and $L$ represents the total number of blocks. The output of the SE-blocks $\mathbf{A}^{4}_{L} \in \mathbb{R}^{T_{a} \times d_{a}}$ then passes through the residual module, where $d_{a}$ represents the output dimension of the SE-block. 

\subsubsection{Residual Module And Regression Module}

The residual module consists of two fully-connected layers (FC), each followed by a layer normalization (LN). It serves two main functions: (1) utilizing the first FC to project speech embeddings into dimensions aligned with linguistic embeddings for cross-modal integration, and (2) leveraging the enriched linguistic information in the integrated embeddings by reintroducing them into the acoustic encoding during CMKT learning. This process is formulated as:
\begin{equation}
\label{eq3}
\begin{aligned}
&\mathbf{E}_{a} = \mathrm{FC_1}(\mathbf{A}^{4}_{L})\\
&\mathbf{A}_{r} = \mathbf{A}^{4}_{L}+ \mathrm{LN}(\mathrm{FC_2}(\mathrm{LN}(\mathbf{E}_{a})))
\end{aligned}
\end{equation}

The speech embedding $\mathbf{E}_{a} \in \mathbb{R}^{T_{a} \times d_{t}}$ is utilized for the CMKT model, where $d_{t}$ represents the dimension of linguistic embedding encoded from the LLM. Subsequently, $\mathbf{A}_{r}$ enters into the regression module to generate a predicted mask $\mathbf{M} \in \mathbb{R}^{T_{a} \times F}$, which is then multiplied back to the noisy spectrogram to produce enhanced spectrogram $\mathbf{\tilde{X}}$. During the inference  stage, the phase information of noisy speech is utilized to reconstruct the enhanced spectrogram into enhanced speech.
\begin{equation}
\label{eq4}
\begin{aligned}
&\mathbf{M} = \mathrm{Sigmoid}(\mathrm{FC}(\mathbf{A}_{r}))\\
&\mathbf{\tilde{X}} = \mathbf{M} \odot \mathbf{X},
\end{aligned}
\end{equation}
where the $\odot$ represent the element-wise multiplication.

\subsection{Cross-Modality Knowledge Transfer learning}
Initially, embeddings containing rich linguistic information are serve as the learning objective for cross-modality alignment. In this study, the LLM is utilized for its exceptional ability to capture semantic and contextual information from text. The linguistic embeddings $\mathbf{\hat{Z}}$ are derived by processing input text through the pre-trained LLM. The process is depicted in the following equation:

\begin{equation}
\label{eq5}
\begin{aligned}
&\mathbf{y}_{token} = \mathrm{Tokenizer}(\mathbf{Y})\\
&\mathbf{\hat{Z}}_{i} = \mathrm{LLM}_{i}(\mathbf{\hat{Z}}_{i-1}),
\end{aligned}
\end{equation}

$\mathrm{LLM}{_i}$ represents the $i$-th encoder layer of the LLM model, where $i$ ranges from $1$ to $M$, with $M$ denoting the total number of LLM encoder layers. Additionally, $\mathbf{\hat{Z}}_{i-1}$ is equal to $\mathbf{y}_{token}$ when $i = 1$. The ‘$\mathrm{Tokenizer}$’ refers to a process that converts standard text into word piece-based tokens. The token $\mathbf{\hat{Z}}_{0}$ is then passed through an embedding layer ($\mathcal{F}_{\mathrm{emb}}$) and augmented with position encoding of text ($\mathrm{PE}_{T}$) to generate textual embeddings $\mathbf{E}_{t}$:

\begin{equation}
\label{eq6}
\begin{aligned}
&\mathbf{y}_{T} = [\mathbf{y}_{\mathrm{BOS}}; \mathbf{y}_{token}; \mathbf{y}_{\mathrm{EOS}}]\\
&\mathbf{E} = \mathcal{F}_{\mathrm{emb}}(\mathbf{y}_{T})\\
&\mathbf{E}_{t} = \mathbf{E}+\mathrm{PE}_{T}(\mathbf{E}),\\
\end{aligned}
\end{equation}
where $\mathbf{y}_{\mathrm{BOS}}$ and $\mathbf{y}_{\mathrm{EOS}}$ are prepended and appended tokens used to implement misalignment strategy, which will be discussed in section \ref{misalign}.

\subsubsection{Cross-Modality Transformer (CMT)}
We utilize the Multi-Head Cross-Attention (MHCA) mechanism in Transformer \cite{vaswani2017attention} to integrate the information of speech embeddings $\mathbf{E}_{a}$ and textual embeddings $\mathbf{E}_{t}$. The cross-modality embedding $\mathbf{Z}$, single-head cross-attention and attention weight $\mathbf{W}_{att}$ can be estimated as:

\begin{equation}
\label{eq7}
\begin{aligned}
&\mathbf{Z} = \mathrm{FFN}(\mathbf{E}_{t}+\mathbf{E}_{\mathrm{CA}})\\
&\mathbf{E}_{\mathrm{CA}} = \mathcal{F}_{\mathrm{CA}}(Q,K,V) = \mathbf{W}_{att} \cdot V\\
&\mathbf{W}_{att} = \mathrm{Softmax}(\frac{QK^{T}}{\sqrt{d_k}}),
\end{aligned}
\end{equation}
where $\mathrm{FFN}$ and $\mathcal{F}_{\mathrm{CA}}$ are the feed-forward networks and the cross-attention operation in Transformer. $K$, $V$ represent the linear transformations of speech embedding $\mathbf{E}_{a}$ by weight $W^{K}$ and $W^{V}$, respectively, while $Q$ denotes the linear transformations of textual embedding $\mathbf{E}_{t}$ by weight $W^{Q}$. And $d_{k}$ is the dimension of $Q$, $K$ and $V$. We enable the cross-modality embedding $\mathbf{Z}$ to approximate the target linguistic embedding from LLM. During the optimization process, linguistic information is encoded in the speech embedding to facilitate knowledge transfer from the LLM to the SE model.

\begin{figure}
\centerline{\includegraphics[width=0.7\columnwidth]{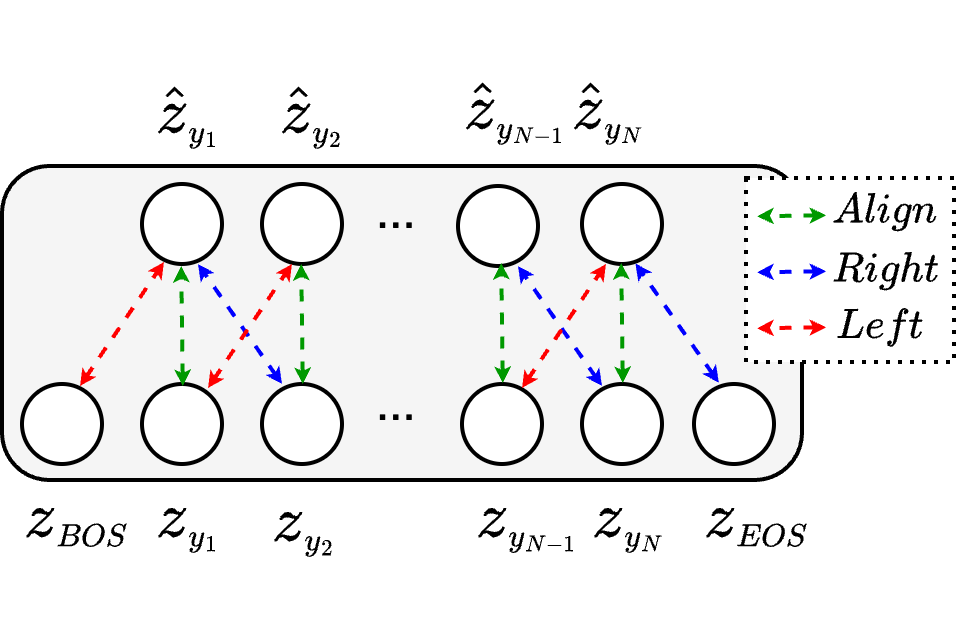}}
\caption{Illustration of the misalignment strategy. The green dashed line, blue dashed line, and red dashed line represent the embeddings being aligned, right-shifted, or left-shifted during the cross-modal alignment loss calculation.
}
\label{fig:misalign}
\end{figure}

\subsubsection{Misalignment Strategy}
\label{misalign}
To enrich the linguistic information within the SE model, a simple yet effective misalignment strategy is introduced into the CMKT learning framework. Compared to aligned linguistic embeddings, shifting the embeddings increases difficulty to the prediction task. This encourages the incorporation of more contextual knowledge into the speech embeddings for better prediction.

Given that the SE model used in this study is non-causal, we explore two types of shifts: right-shifted (blue dashed line) and left-shifted (red dashed line), as illustrated in Fig. \ref{fig:misalign}. During the extraction of cross-modality embeddings, both $\mathbf{y}_{\mathrm{BOS}}$ and $\mathbf{y}_{\mathrm{EOS}}$ tokens are added to the input, as shown in Eq. (\ref{eq6}). For loss computation, the embeddings are shifted in different directions to simulate the misaligned conditions. To ensure a fair comparison of the effects of this strategy, the aligned and misaligned configurations are designed to have identical trainable parameters and computational costs.

\subsection{Loss Function}
The total loss function composed as a linear combination of the SE loss and cross-modal alignment loss ($\mathcal{L}_{\mathrm{CMA}}$). The SE loss is calculated as the mean absolute error ($\mathcal{L}_{\mathrm{MAE}}$) between the enhanced spectrogram and the clean spectrogram, while the CMA loss endeavors to maximize the cosine similarity between cross-modality embeddings ($\mathbf{Z}$) and context-dependent linguistic embeddings ($\mathbf{\hat{Z}}$) from LLMs: 
\begin{equation}
\label{eq8}
\begin{aligned}
&\mathcal{L}_{\mathrm{Total}} = 
\alpha\cdot\mathcal{L}_{\mathrm{MAE}}
+(1-\alpha)\cdot\mathcal{L}_{\mathrm{CMA}},\\
&\mathcal{L}_{\mathrm{MAE}} =\mathbb{E}\big[\lVert \mathbf{\hat{X}}-\mathbf{\tilde{X}} \rVert^{1}_{1} \big],\\
&\mathcal{L}_{\mathrm{CMA}} =\sum\nolimits_{t=1}^{T_{t}}(1-\cos(\mathbf{z}_{t},\mathbf{\hat{z}}_{t})),
\end{aligned}
\end{equation}
where $\alpha$ is the hyper-parameter to control the weighting of the two losses. $T_{t}$ is the length of the cross-modality embeddings. Additionally, $\mathbf{\tilde{X}}$ represents the enhanced spectrogram, while $\mathbf{\hat{X}}$ denotes the clean spectrogram. Finally, $\mathbf{z}_{t}$ and $\mathbf{\hat{z}}_{t}$ are vectors of matrices $\mathbf{Z}$ and $\mathbf{\hat{Z}}$, respectively.

\section{Experiments}
\label{sec:experiments}

The proposed CMKT learning is designed as a generalized approach for SE. To evaluate its effectiveness, our primary experiments were conducted on a Mandarin speech dataset. Additionally, we extended the evaluation to an English speech dataset to verify the applicability of the method across different languages. Furthermore, scenarios where training data lacked the the corresponding text were investigated to assess the practicality and robustness of the proposed method. The following sections provide a detailed description of the experimental setup, present the results, and provide a comprehensive analysis and discussion of the findings.

\subsection{Experiments Setup}
This study investigates the effectiveness of transferring linguistic knowledge from a LLM to a SE model within the proposed CMKT learning framework. The experiments were conducted using various combinations of SE models and LLMs for the comprehensive assessment. The baseline approach corresponds to the left branch of the architecture depicted in Fig. \ref{fig:model}.

\subsubsection{Model}
The proposed method consists of two branches: the SE model and the CMKT module. For the SE model, in addition to the Conformer block mentioned in Section \ref{sec:seblock}, we also employed Transformer \cite{vaswani2017attention} and bidirectional long short-term memory (BLSTM) \cite{graves2005framewise} blocks as variants of SE-blocks to further evaluate the performance across different architectures. The SE model processes log amplitude spectrograms as speech input, which are computed using short-time Fourier transform (STFT) with a Hamming window. The window length is set to 25 ms, and the hop size is 6.25 ms. The dimensions of the speech features ($d_{a}$) and linguistic embeddings ($d_{t}$) are set to 201 and 768, respectively. The CNN encoder consists of two convolutional layers, each with a kernel size of 3 and a stride of 1. Within each Conformer block, the convolution operation uses a kernel size of 15, while the attention dimension, attention heads, and feed-forward network (FFN) layer dimensions are configured as 256, 4, and 2048, respectively. The number of SE-blocks is configured as 4 for both Conformer and Transformer blocks, and 5 for BLSTM blocks.

For the CMKT learning, the CMT module consists of three Transformer encoder layers, in which the self-attention is replaced with cross-attention. The attention dimension, number of attention heads, and FFN layer dimension are set to 768, 12, and 2048, respectively. The model is optimized using the Adam optimizer \cite{kingma2014adam} with an initial learning rate of 0.001, employing a warm-up schedule of 20,000 steps. Training is conducted over 130 epochs, with the final evaluation model obtained by averaging the parameters from the last 10 epochs. The weighting factor $\alpha$ in the Eq. (\ref{eq8}) is fixed at 0.7.

\subsubsection{LLMs}
In our experiments, we employed three widely adopted large language models (LLMs): BERT \cite{devlin2018bert}, LLaMA-2 \cite{touvron2023llama}, and Alpaca \cite{taori2023alpaca}. These models are commonly used in open-source applications and represent distinct methodologies in language modeling. By leveraging these well-established models, we aim to evaluate the effectiveness of our proposed method under diverse linguistic embedding strategies.

BERT is pre-trained with masked language modeling (MLM) and next sentence prediction (NSP), enabling it to capture context-sensitive embeddings that reflect both local and global semantic relationships. LLaMA-2 builds upon its predecessor, LLaMA \cite{Touvron2023LLaMAOA}, with several architectural enhancements, including rotary positional embeddings (RoPE) and grouped-query attention (GQA), which improve computational efficiency and scalability. Moreover, LLaMA-2 is trained on a larger and more diverse dataset, resulting in stronger generalization capabilities. Alpaca, a fine-tuned variant of the original LLaMA, specializes in instruction-following tasks, providing a lightweight and efficient solution tailored for practical downstream applications.

BERT utilizes an encoder-based design, while LLaMA-2 and Alpaca adopt decoder-based approaches. Despite these architectural differences, all three models are built upon large-scale pretraining on diverse datasets. These models were chosen for their widespread recognition and to evaluate the adaptability of the proposed method across different LLM architectures. This selection highlights the method's robustness and applicability across varying linguistic embeddings.

\begin{table*}[!htbp]
    \caption{Evaluation results under different SNR levels on the AISHELL-1 dataset. Various combinations of model architectures and LLM embeddings were tested. The bold values indicate the best results within the same model architecture.}
    \label{table:score}
    \small
    \setlength{\tabcolsep}{3pt}
    \centering
    \begin{tabular}{l||ccc|ccc|ccc|ccc}
    \toprule[0.4mm]
    \multirow{2}[2]{*}{Method} &  
    \multicolumn{3}{c}{High SNR} &
    \multicolumn{3}{c}{Medium SNR} &
    \multicolumn{3}{c}{Low SNR} &
    \multicolumn{3}{c}{Avg.} \\
 \cmidrule(lr){2-4} \cmidrule(lr){5-7} \cmidrule(lr){8-10} \cmidrule(lr){11-13}
 & PESQ & STOI& VQScore& PESQ & STOI& VQScore&PESQ & STOI& VQScore&PESQ & STOI& VQScore\\
    \midrule[0.4mm]
    Noisy&  1.93 & 0.887 & 0.643 & 1.264 & 0.74 & 0.606 & 1.073 & 0.507 & 0.578 & 1.439 & 0.718 & 0.61\\
    \midrule[0.2mm]

    Conformer &2.668 & 0.913 & 0.663 & 1.723 & 0.796 & 0.641 & 1.224 & 0.579 & 0.619 & 1.898 & 0.768 & 0.642\\
    \quad +CMKT (BERT)& \textbf{2.911} & \textbf{0.928} & \textbf{0.676} & \textbf{2.01} & \textbf{0.835} & \textbf{0.665} & \textbf{1.333} & \textbf{0.623} & \textbf{0.649} & \textbf{2.113} & \textbf{0.8} & \textbf{0.664} \\
    \quad +CMKT (LLaMA)& 2.903 & 0.925 & 0.674 & 1.93 & 0.821 & 0.659 & 1.291 & 0.604 & 0.641 & 2.098 & 0.796 & 0.662\\
    \quad +CMKT (Alpaca)& 2.894 & 0.926 & 0.675 & 1.977 & 0.829 & 0.662 & 1.312 & 0.617 & 0.645 & 2.09 & 0.796 & 0.661 \\
    \midrule[0.2mm]
    Transformer & 2.682 & 0.913 & 0.664 & 1.708 & 0.791 & 0.643 & 1.217 & 0.563 & 0.618 & 1.896 & 0.762 & 0.642\\
    \quad +CMKT (BERT) & \textbf{2.908} & \textbf{0.925} & \textbf{0.676} & \textbf{2.001} & \textbf{0.83} & \textbf{0.664} & \textbf{1.327} & \textbf{0.613} & \textbf{0.643} & \textbf{2.107} & \textbf{0.794} & \textbf{0.662}\\
    \quad +CMKT (LLaMA) & 2.838 & 0.919 & 0.673 & 1.866 & 0.807 & 0.656 & 1.259 & 0.579 & 0.635 & 2.016 & 0.774 & 0.655\\
    \quad +CMKT (Alpaca) & 2.825 & 0.919 & 0.672 & 1.854 & 0.806 & 0.655 & 1.254 & 0.575 & 0.631 & 2.006 & 0.772 & 0.654\\
    \midrule[0.2mm]
    BLSTM & 2.781 & 0.918 & 0.666 & 1.81 & 0.805 & 0.646 & 1.251 & 0.585 & 0.624 & 1.975 & 0.775 & 0.646\\
    \quad +CMKT (BERT) & 2.896 & 0.927 & \textbf{0.679} & \textbf{2.047} & \textbf{0.841} & \textbf{0.674} & \textbf{1.373} & \textbf{0.653} & \textbf{0.658} & \textbf{2.133} & \textbf{0.811} & \textbf{0.671}\\
    \quad +CMKT (LLaMA) & \textbf{2.952} & \textbf{0.928} & 0.676 & 2.004 & 0.832 & 0.665 & 1.324 & 0.622 & 0.648 & 2.123 & 0.799 & 0.663 \\
    \quad +CMKT (Alpaca) & 2.941 & 0.927 & 0.675 & 2.005 & 0.829 & 0.662 & 1.325 & 0.616 & 0.645 & 2.12 & 0.796 & 0.661\\

    \bottomrule[0.4mm]
    \end{tabular}%
\end{table*}

\subsubsection{Evaluation Metrices}\label{evaluation}
The evaluation metrics comprise three key standards: perceptual evaluation of speech quality (PESQ) \cite{rix2001perceptual}, short-time objective intelligibility (STOI) \cite{taal2011algorithm}, and VQScore \cite{fu2024self}. PESQ and STOI serve as prevalent measures for assessing speech quality and intelligibility, respectively. Meanwhile, VQScore is specifically employed to evaluate the mean opinion score (MOS) and demonstrates remarkable robustness, notably evident in its effectiveness when applied to out-of-domain data, owing to its unsupervised approach. This combination of metrics forms a comprehensive evaluation framework, ensuring an in-depth assessment of speech enhancement performance under diverse conditions.

\subsection{Experiments on Mandarin Speech Corpus}
Experiments were conducted using the open-source Mandarin speech corpus, AISHELL-1, comprising speech and text data from 400 speakers \cite{bu2017aishell}. The corpus consists of three distinct datasets: a training set encompassing 340 speakers (150 hours), a development set with 40 speakers (10 hours), and a test set comprising 20 speakers (5 hours). For the introduction of noise, we employed the noise dataset provided in this study \cite{lin2021unsupervised}, following a methodology that involved partitioning the noise dataset into two subsets. 
Signal-to-noise ratios (SNRs) spanning from -15 to 15 were randomly selected for synthesizing both the training and test datasets, ensuring comprehensive assessment across varied noise levels.

\begin{figure*}[!htbp]
\centerline{\includegraphics[width=2\columnwidth]{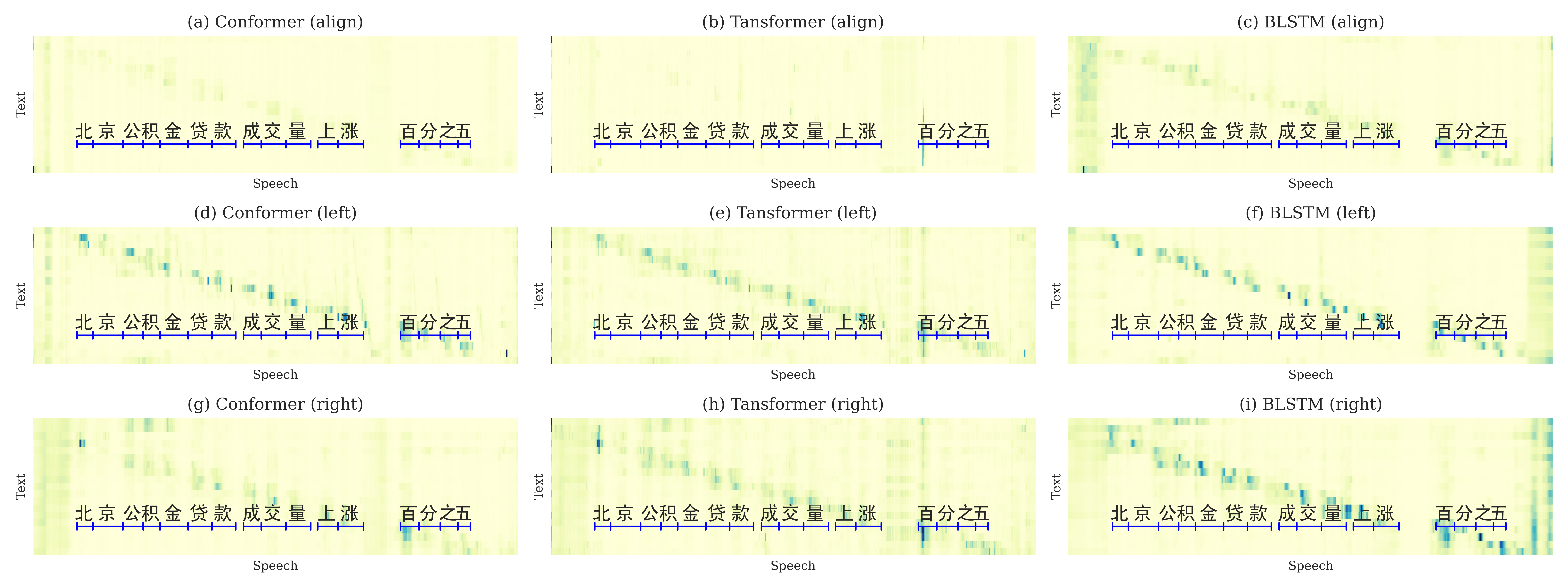}}
\caption{The attention weighting in the cross-modality module, where the x-axis and y-axis represent the speech and text modalities, respectively. Results for different models under various alignment methods are presented: (a)–(c) correspond to aligned, (d)–(f) to left-shifted, and (g)–(i) to right-shifted alignments.}
\label{fig:attention}
\end{figure*}

\begin{table}[!htbp]
    \caption{Evaluation results for different alignment methods on the AISHELL-1 dataset. The bold values indicate the best results within the same model architecture.}
    \label{table:alignment}
    \small
    \setlength{\tabcolsep}{3pt}
    \centering
    \begin{tabular}{lccccc}
    \toprule[0.4mm]
    \multirow{2}[2]{*}{Method} & 
    \multirow{2}[2]{*}{Alignment} &
    \multicolumn{3}{c}{Evaluation} \\
 \cmidrule(lr){3-5} 
 && PESQ & STOI& VQScore\\
    \midrule[0.2mm]

    \multirow{3}{*}{Conformer} & Align & 2.052 & 0.786 & 0.656 \\
    & Right & 2.087 & \textbf{0.8} & 0.663 \\
    & Left & \textbf{2.113} & \textbf{0.8} & \textbf{0.664} \\
    \midrule[0.2mm]
    \multirow{3}{*}{Transformer} & Align & 2.037 & 0.778 & 0.656 \\
    & Right & \textbf{2.117} & 0.793 & 0.661 \\
    & Left & 2.107 & \textbf{0.794} & \textbf{0.662} \\
    \midrule[0.2mm]
    \multirow{3}{*}{BLSTM} & Align & 2.102 & 0.794 & 0.661  \\
    & Right & 2.113 & 0.803 & 0.668 \\    
    & Left & \textbf{2.133} & \textbf{0.811} & \textbf{0.671}\\
    \bottomrule[0.4mm]
    \end{tabular}%
\end{table}

The pre-trained LLMs used in our experiments—BERT, LLaMA-2, and Alpaca—are open-source models with publicly available weights obtained from HuggingFace. The Chinese BERT\footnote{\url{https://huggingface.co/google-bert/bert-base-chinese}} was pre-trained on a large-scale Chinese corpus, making it adept at capturing linguistic nuances. The Chinese LLaMA-2\footnote{\url{https://huggingface.co/hfl/chinese-llama-2-7b}} builds on the original LLaMA-2 architecture with an additional 120GB of Chinese text data, enhancing its capacity to understand and generate Chinese text. The Chinese Alpaca\footnote{\url{https://huggingface.co/hfl/chinese-alpaca-2-7b}}, fine-tuned from the Chinese LLaMA-2, further incorporates approximately 5 million instruction-following examples, optimizing it for instruction-based tasks. These models ensure experimental consistency, reproducibility, and a robust evaluation of the proposed method across diverse linguistic embeddings.

\subsubsection{Result for Speech Enhancement}\label{cmkt}
Table \ref{table:score} summarizes the evaluation results of the proposed CMKT learning under three different SNR levels: high (5 to 15 dB), medium (-5 to 5 dB), and low (-15 to -5 dB). These levels correspond to mild, moderate, and severe noise conditions, ensuring a comprehensive assessment. The results demonstrate that CMKT consistently improves the performance of all baseline SE models across PESQ, STOI, and VQScore metrics, confirming the effectiveness of leveraging LLM embeddings in SE.

Among the evaluated LLM embeddings, BERT achieves slightly higher average scores compared to LLaMA and Alpaca. For instance, in the Transformer model, CMKT (BERT) achieves an average PESQ of 2.107, which is slightly higher than CMKT (LLaMA) and CMKT (Alpaca), both scoring around 2.02. Similarly, CMKT (BERT) attains the highest STOI score of 0.794, with CMKT (LLaMA) and CMKT (Alpaca) scoring 0.784 and 0.782, respectively. On the other hand, for the Conformer and BLSTM models, the differences among the three LLM embeddings are relatively minor. These results indicate that while the choice of LLM embedding may influence performance more significantly in certain SE architectures, all embeddings effectively enhance the baseline models, consistently improves performance across all tested configurations. Overall, the proposed CMKT method demonstrates robust performance enhancements across all tested configurations, with BERT embeddings delivering slightly better results. Consequently, we select BERT as the primary configuration for subsequent analyses.

\subsubsection{Effectiveness of Misalignment Strategy}\label{misalignment}
Table \ref{table:alignment} summarizes the evaluation results of the proposed misalignment strategy under three alignment methods: aligned, left-shifted (left), and right-shifted (right). Both misalignment strategies (left and right) consistently outperform the aligned method across all SE models, demonstrating their effectiveness in enhancing SE performance. Among the two misalignment strategies, the left-shifted alignment achieves slightly better results than the right-shifted alignment in most cases, highlighting its potential as a robust approach for linguistic knowledge transfer.

The attention weighting in the cross-modality module, which could be calculated from Eq. (\ref{eq7}), is visualized in Fig. \ref{fig:attention}, It provides deeper insights into the impact of the misalignment strategy. The attention maps illustrate the relationship between speech and text embeddings, with the x-axis representing the frame indices of speech embeddings and the y-axis corresponding to the indices of text tokens. To enhance interpretability, we provided word-level force alignment using the Charsiu framework\footnote{https://github.com/lingjzhu/charsiu}, and the aligned positions are marked on the attention maps for reference. In particular, this alignment information was used exclusively for visualization purposes, and the model does not rely on explicit alignment data during training.

From the attention maps, two significant observations can be made. First, the model demonstrates a clear ability to associate text and audio embeddings effectively. For example, speech silence regions exhibit minimal attention weights, while for non-silent regions, the attention weights align closely with the reference word alignments we provided. This behavior reflects the model's capacity to focus on relevant cross-modality information during the learning process. Second, the diagonal patterns observed in the attention maps highlight the quality of the learned alignment between text and audio embeddings. In particular, the left- and right-shifted configurations exhibit a more distinct diagonal structure compared to the aligned setup. This suggests that the misalignment strategy promotes better generalization in learning linguistic relationships, enabling the model to capture more precise correlations between modalities. These findings explain the superior performance of misaligned strategies and underscore their importance in facilitating effective linguistic knowledge transfer within the CMKT framework.

\begin{figure*}
\centerline{\includegraphics[width=2\columnwidth]{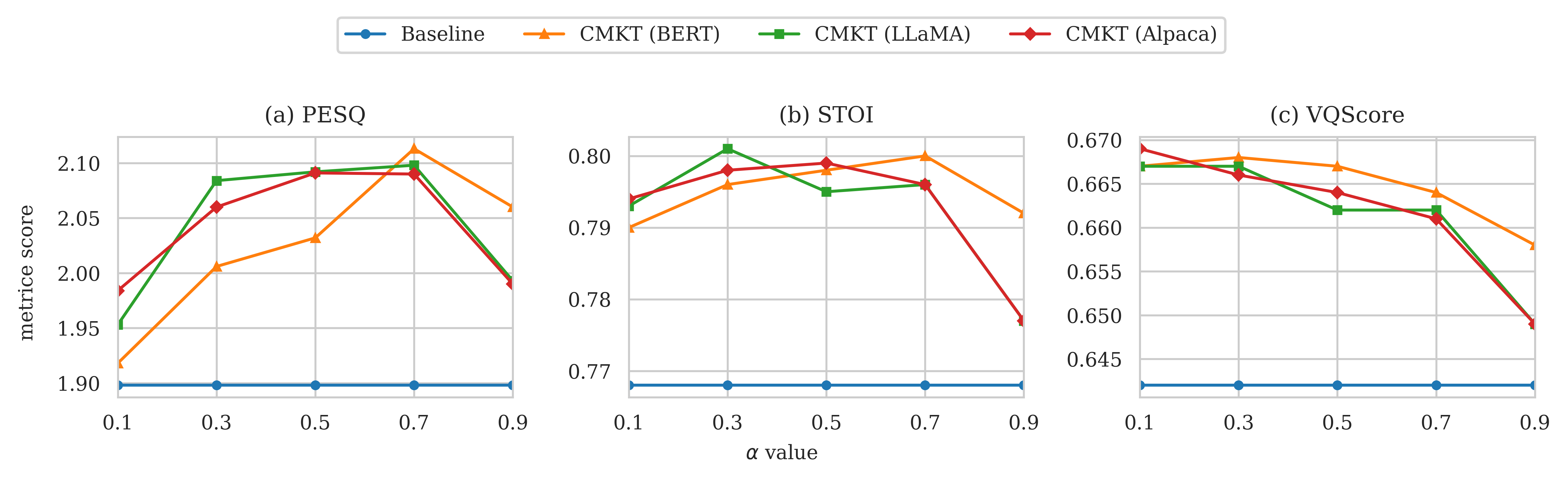}}
\caption{The evaluation scores at different $\alpha$ values (0.1 to 0.9 in increments of 0.2) on the AISHELL-1 dataset. The blue line represents the baseline Conformer model without CMKT.}
\label{fig:alpha_diff}
\end{figure*}

\subsubsection{Investigation on different $\alpha$ Values}\label{alpha}
The parameter $\alpha$, defined in Eq. (\ref{eq8}), determines the extent to which linguistic information is integrated into the acoustic embeddings. To evaluate the sensitivity of the proposed approach to different $\alpha$ values, we tested a range of values from 0.1 to 0.9 in increments of 0.2, as illustrated in Fig. \ref{fig:alpha_diff}. A lower $\alpha$ value assigns greater importance to CMKT learning, allowing more linguistic information to be integrated into the acoustic embeddings. Regardless of the $\alpha$ value, all configurations incorporating CMKT outperform the baseline Conformer model (blue line), demonstrating the consistent advantages of the proposed framework.

For PESQ, performance steadily improves as $\alpha$ increases, reaching its peak at $\alpha=0.7$. In contrast, STOI remains relatively stable throughout the range of $\alpha$ values. For VQScore, an negative correlation is observed, with scores gradually decreasing as $\alpha$ increases. This decline becomes more pronounced between $\alpha=0.7$ and $\alpha=0.9$, suggesting that higher $\alpha$ values might limit the model's ability to effectively balance linguistic and acoustic features. Additionally, since Alpaca is fine-tuned from LLaMA, their results exhibit similar trends across all metrics, reflecting the close relationship between the two models and the impact of their initial pre-training on downstream tasks.

\begin{figure*}[!tb]
\centerline{\includegraphics[width=2\columnwidth]{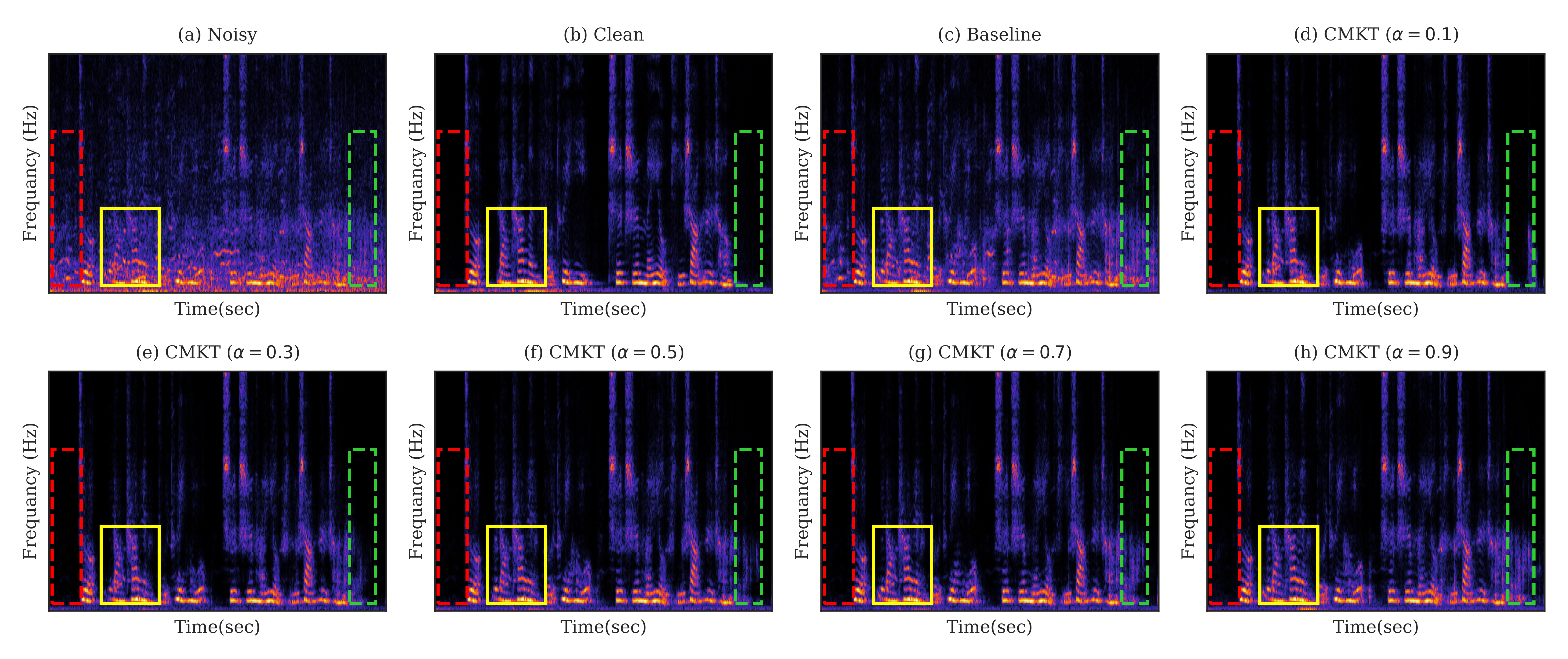}}
\caption{Spectrogram plots under different methods: (a) Noisy, (b) Clean, (c) Baseline (Conformer), and (d)–(h) CMKT with different $\alpha$ values.}
\label{fig:spec}
\end{figure*}

In addition to quantitative evaluation, we provide the magnitude spectrograms for a test utterance in Fig. \ref{fig:spec}. Three regions of interest are highlighted in the spectrograms: two noise-only segments (red and green dashed box) and one mixed speech-noise regions (yellow box). These visualizations offer insights into how varying $\alpha$ values affect the trade-off between denoising effectiveness and speech fidelity.

In the first noise-only regions (red dashed box), all CMKT configurations show significant improvements over the baseline by effectively reducing noise. The noise suppression in these regions is consistent across all $\alpha$ values, demonstrating the robustness of CMKT in eliminating non-speech noise. In the second noise-only regions (green dashed box), higher $\alpha$ values further enhance noise removal, leading to cleaner spectrograms. However, in the mixed speech-noise regions (yellow box), smaller $\alpha$ values introduce more noticeable distortions to the reconstructed speech. This trade-off highlights that while lower $\alpha$ values improve noise suppression, they may compromise the preservation of speech-specific features, affecting overall reconstruction quality. These observations emphasize the importance of selecting an appropriate $\alpha$ value to balance linguistic knowledge integration and acoustic feature preservation.

\subsection{Experiments on English Speech Corpus}
The experiments were conducted using the publicly available English speech corpus, LibriSpeech, which comprises read speech recordings derived from audiobooks in the LibriVox project \cite{panayotov2015librispeech}. The "train-clean-360" subset, containing 360 hours of speech from 921 speakers, was used for training. For evaluation, we employed the "test-clean" subset, which includes 5.4 hours of speech from 40 distinct speakers. The training and test sets were designed with no speaker overlap, ensuring that the evaluation accurately measures the model's ability to generalize to unseen data. For noise augmentation, we employed the same noise dataset used in the Mandarin experiments. The noise data were divided into training and testing subsets to prevent overlap. 
Signal-to-noise ratios (SNRs) were randomly sampled between -15 dB and 15 dB to synthesize noisy speech for both training and testing, enabling a comprehensive evaluation of the model's robustness under diverse noise conditions.

We adopted the same LLM architectures as in the Mandarin experiments: BERT, LLaMA-2, and Alpaca, all obtained from HuggingFace. The English BERT\footnote{\url{https://huggingface.co/google-bert/bert-base-uncased}} was pre-trained on a large-scale English text corpus, enabling it to capture contextual and semantic relationships effectively. LLaMA-2\footnote{\url{https://huggingface.co/meta-llama/Llama-2-7b-hf}} was trained on over 2 trillion tokens of English data, providing extensive language coverage. Alpaca\footnote{\url{https://huggingface.co/wxjiao/alpaca-7b}} was fine-tuned from the original LLaMA model in two stages: the first stage involved general-purpose fine-tuning, while the second stage incorporated approximately 52,000 instruction-following examples to enhance task-specific performance. This experiment broadens our evaluation to a different linguistic domain, offering deeper insights into the CMKT framework’s effectiveness across languages.

Table \ref{table:english} presents the evaluation results on the English LibriSpeech dataset for three SE model architectures combined with different CMKT configurations. Consistent with the findings from the Mandarin dataset, integrating CMKT with pre-trained LLM embeddings leads to significant performance improvements across all models and metrics compared to the baseline SE models without CMKT.

The best-performing LLM varies across SE models, with CMKT (Alpaca) achieving the highest scores for Conformer and BLSTM, while CMKT (LLaMA) performs best for Transformer. Comparing the results with the AISHELLL-1 dataset (Mandarin) in Table \ref{table:score}, we observe consistent trends across both languages. CMKT enhances performance regardless of the baseline SE model's initial capabilities, demonstrating its robustness and adaptability. For instance, the BLSTM baseline already outperforms Transformer combined with CMKT (LLaMA), yet the integration of CMKT further boosts BLSTM’s performance. These findings validate the generalizability of the CMKT framework across different languages and architectures, underscoring its potential for cross-lingual SE applications.

\begin{table}[!tb]
    \caption{Evaluation scores on the Librispeech dataset (English). The bold values indicate the best scores within the same architecture.}
    \label{table:english}
    \small
    \setlength{\tabcolsep}{3pt}
    \centering
    \begin{tabular}{lccc}
    \toprule[0.4mm]
    \multirow{2}[2]{*}{Method} & 
    \multicolumn{3}{c}{Evaluation} \\
 \cmidrule(lr){2-4} 
 & PESQ & STOI& VQScore\\
    \midrule[0.4mm]
    Noisy & 1.398 & 0.808 & 0.623 \\
    \midrule[0.2mm]
    Conformer & 1.664 & 0.854 & 0.664 \\
    \quad +CMKT (BERT) & 1.715 & 0.858 & 0.673 \\
    \quad +CMKT (LLaMA) & 1.726 & 0.858 & 0.674 \\
    \quad +CMKT (Alpaca) & \textbf{1.747} & \textbf{0.859} & \textbf{0.676} \\
    \midrule[0.2mm]
    Transformer  & 1.658 & 0.844 & 0.666 \\
    \quad +CMKT (BERT) & 1.694 & 0.849 & 0.671 \\
    \quad +CMKT (LLaMA) & \textbf{1.718} & \textbf{0.847} & \textbf{0.673} \\
    \quad +CMKT (Alpaca) & 1.68 & 0.846 & 0.668 \\
    \midrule[0.2mm]
    BLSTM  & 1.738 & 0.853 & 0.673\\
    \quad +CMKT (BERT)  & 1.769 & 0.854 & \textbf{0.683} \\
    \quad +CMKT (LLaMA) & 1.768 & \textbf{0.856} & 0.682 \\
    \quad +CMKT (Alpaca) & \textbf{1.789} & \textbf{0.856} & 0.682 \\
    \bottomrule[0.4mm]
    \end{tabular}%
\end{table}

\subsection{Experiment with ASR transcription}
To explore the impact of CMKT in scenarios where corresponding text is unavailable, we employed two commonly used multilingual ASR models: XLSR-53 \cite{conneau2020unsupervised} and Whisper \cite{radford2023robust}. The XLSR-53 model, pre-trained on 53 languages using a contrastive loss on masked speech representations, leverages cross-lingual shared representations to enhance performance across diverse languages. Meanwhile, the Whisper model, trained on a massive dataset 680,000 hours encompassing multilingual and multitask supervision, exhibits strong zero-shot transfer capabilities without the need for fine-tuning. Pre-trained weights for XLSR-53\footnote{\url{https://huggingface.co/jonatasgrosman/wav2vec2-large-xlsr-53-chinese-zh-cn}} and Whisper\footnote{\url{https://huggingface.co/openai/whisper-large-v3-turbo}} were obtained from HuggingFace.

The experimental setup was consistent with the methodology established in the previous experiments. In this study, we employed distinct SE architectures and integrated BERT embeddings to facilitate CMKT learning. During the speech recognition process, we observed instances where the ASR model failed to generate outputs. For such cases, the corresponding speech segments were trained solely using the SE loss. Notably, the parameter $\alpha$ in Eq. (\ref{eq8}) was fixed at 0.7, regardless of whether text data were available, to ensure a fair comparison with previous experiments.

\begin{table}[!tb]
    \caption{ASR performance with different ASR models on the AISHELL-1 dataset.}
    \label{table:cer}
    \small
    \setlength{\tabcolsep}{3pt}
    \centering
    \begin{tabular}{l|ccc}
    \toprule[0.4mm]
    \multirow{2}[2]{*}{ASR model} &
    \multicolumn{3}{c}{CER (\%)}\\
    \cmidrule(lr){2-4} 
    & train & dev & test \\
    \midrule[0.4mm]
    Whisper & 11.64 & 10.29 & 10.41 \\
    \midrule[0.2mm]
    XLSR-53 & 20.28 & 19.03 & 21.22 \\
    \bottomrule[0.4mm]
    \end{tabular}%
\end{table}

\begin{table}[!tb]
    \caption{Evaluation scores with different transcriptions on the AISHELL-1 dataset. The impact of different alignment methods on various ASR systems is also compared. Linguistic embeddings are extracted using the BERT model. The bold values indicate the overall best results.}
    \label{table:asr}
    \small
    \setlength{\tabcolsep}{3pt}
    \centering
    \begin{tabular}{lcccccc}
    \toprule[0.4mm]
    \multirow{2}[2]{*}{Method} & \multirow{2}[2]{*}{Text} & \multirow{2}[2]{*}{Alignment} &
    \multicolumn{3}{c}{Evaluation}\\
 \cmidrule(lr){4-6} 
 &&& PESQ & STOI& VQScore \\
    \midrule[0.4mm]
    Conformer & -- & -- & 1.898 & 0.768 & 0.642 \\
    \midrule[0.2mm]
    \multirow{7}{*}{\quad +CMKT} & Oracle & Left & \textbf{2.113} & \textbf{0.8} & \textbf{0.664} \\
    \cmidrule(lr){2-6}
    & \multirow{3}{*}{Whisper} & Align & 2.067 & 0.788 & 0.657 \\
    && Right & 2.101 & 0.799 & \textbf{0.664} \\
    && Left & 2.091 & 0.799 & \textbf{0.664} \\
    \cmidrule(lr){2-6}
    & \multirow{3}{*}{XLSR-53} & Align & 2.059 & 0.787 & 0.657 \\
    && Right & 2.082 & 0.798 & 0.663 \\
    && Left & 2.085 & 0.8 & 0.663 \\
    \midrule[0.4mm]
    Transformer & -- & -- & 1.896 & 0.762 & 0.642 \\
    \midrule[0.2mm]
    \multirow{7}{*}{\quad +CMKT} & Oracle & Left & \textbf{2.107} & \textbf{0.794} & \textbf{0.662} \\
    \cmidrule(lr){2-6}
    & \multirow{3}{*}{Whisper} & Align & 2.022 & 0.78 & 0.656 \\
    && Right & 2.105 & 0.793 & 0.661 \\
    && Left & 2.106 & \textbf{0.794} &  \textbf{0.662} \\
    \cmidrule(lr){2-6}
    & \multirow{3}{*}{XLSR-53} & Align & 2.024 & 0.78 & 0.655 \\
    && Right & 2.082 & 0.792 & 0.661 \\
    && Left & 2.072 & 0.791 & 0.661 \\
    \midrule[0.4mm]
    BLSTM & -- & -- & 1.975 & 0.775 & 0.646 \\
    \midrule[0.2mm]
    \multirow{7}{*}{\quad +CMKT} & Oracle & Left & \textbf{2.133} & \textbf{0.811} & \textbf{0.671} \\
    \cmidrule(lr){2-6}
    & \multirow{3}{*}{Whisper} & Align & 2.095 & 0.805 & 0.67 \\
    && Right & 2.132 & 0.806 & \textbf{0.671} \\
    && Left & 2.106 & 0.807 & 0.67 \\
    \cmidrule(lr){2-6}
    & \multirow{3}{*}{XLSR-53} & Align & 2.094 & 0.794 & 0.66 \\
    && Right & 2.123 & 0.798 & 0.663 \\
    && Left & 2.117 & 0.796 & 0.661 \\
    \bottomrule[0.4mm]
    \end{tabular}%
\end{table}

\subsubsection{Result for Speech Enhancement}
Table \ref{table:cer} reports the character error rate (CER) performances of the ASR models. Whisper demonstrates significantly higher transcription accuracy compared to XLSR-53 on the AISHELL-1 dataset, with consistent CERs across the train, dev, and test sets. These results establish a foundation for evaluating the impact of transcription quality on CMKT performance in speech enhancement.

Table \ref{table:asr} illustrates the evaluation results, comparing the performance of CMKT using oracle (ground truth text), Whisper, and XLSR-53 transcriptions across different SE models. CMKT consistently improves the baseline SE models across all configurations, irrespective of the transcription source. Higher transcription accuracy leads to better enhancement performance, as seen in Whisper’s results, which outperform XLSR-53 across all SE architectures. Even with the less accurate transcriptions from XLSR-53, CMKT still provides substantial gains over the baseline models, demonstrating its robustness in leveraging linguistic embeddings effectively. While Whisper does not fully match the performance of oracle transcriptions, it achieves results that are indistinguishable from oracle in certain metrics. For instance, in the Transformer model, Whisper achieves a STOI of 0.794, matching Oracle’s performance. Similarly, in the BLSTM model, Whisper attains a VQScore of 0.671, identical to Oracle. These findings highlight the effectiveness of CMKT even with non-ideal transcriptions, reinforcing its applicability in real-world SE tasks.

\subsubsection{Effectiveness of Misalignment Strategy}
The misalignment strategy is shown to effectively enhance CMKT performance across all transcription methods and SE architectures. The performance differences between the left-shifted and right-shifted alignments are minimal. However, both consistently outperform the aligned configuration, highlighting the benefits of introducing temporal shifts when integrating linguistic embeddings. For instance, in the Conformer model with Whisper transcriptions, both left-shifted and right-shifted alignments achieve a VQScore of 0.664, compared to 0.657 for the aligned configuration. Similarly, in the Transformer model with XLSR-53 transcriptions, the right-shifted alignment improves PESQ from 2.024 (aligned) to 2.082, and STOI from 0.78 to 0.792. These consistent improvements demonstrate the robustness of the misalignment strategy across different transcription qualities and SE architectures, further enhancing the adaptability of CMKT.

These findings underscore the importance of transcription quality and alignment strategies in optimizing CMKT performance. Whisper, with its superior transcription accuracy, consistently enables CMKT to achieve results that closely approach those obtained with oracle transcriptions, particularly when combined with the misalignment strategy. Notably, the misalignment strategy proves effective across diverse configurations, consistently outperforming the aligned configuration. Furthermore, CMKT demonstrates substantial improvements over baseline models across all SE architectures and transcription sources, highlighting its robustness and adaptability for real-world SE applications under varying transcription conditions.

\section{Conclusion}
\label{sec:conclusion}
This study introduced the CMKT learning framework, which effectively transfers linguistic knowledge from pre-trained LLMs into SE models. The proposed approach eliminates the need for text input or LLM involvement during inference, making it a practical and scalable solution for real-world applications. Experimental results confirmed that CMKT consistently enhances SE performance across various architectures and different LLMs, confirming its robustness and adaptability. Additionally, we introduced a misalignment strategy that incorporates temporal shifts to optimize the integration of linguistic and acoustic representations. This strategy consistently outperforms aligned configurations, leading to improved SE performance across different SE models. 

The framework was evaluated on both Mandarin and English datasets, yielding consistent performance improvements across different language corpora. Furthermore, experiments conducted without corresponding text data confirmed that CMKT remains effective even with automatically generated transcriptions, demonstrating its robustness in handling real-world scenarios with limited or noisy linguistic resources. In future work, we plan to investigate the influence of linguistic representations from different LLM layers to better understand their contribution to SE performance. Additionally, we aim to expand CMKT to other speech-related tasks to further explore its applicability. These findings establish a solid foundation for bridging linguistic and acoustic domains in SE and advancing cross-modality learning.




\bibliographystyle{IEEEbib}
\bibliography{refs}

\vfill

\end{document}